\documentclass{article}
\usepackage{arxiv}

\usepackage{cite}
\usepackage{amsmath,amssymb,amsfonts}
\usepackage{algorithmic}
\usepackage{multirow}
\usepackage{graphicx}
\usepackage{textcomp}

\usepackage{verbatim} %for block commenting
\usepackage{soul} %for text highlighting
\usepackage{wasysym} %for smileys
\usepackage{todonotes}
\usepackage[T1]{fontenc}
\usepackage[utf8]{inputenc}

\title{Implementing a chatbot solution for learning management systems}

\author{
Dimitrios Chaskopoulos, Jonas Eilertsen Hægdahl, Petter Sagvold, Claire Trinquet, Maryam Edalati 
\\
 \textit{Department of Computer Science} \\
    \textit{Norwegian University of Science and Technology}\\
    Gjøvik, Norway \\
    \{dimitrc, jonaseha, pettsag, clairet, maryame\}@stud.ntnu.no
}

\begin{document}
\maketitle

\begin{abstract}
Innovation is a key component in trying new solutions for the students to learn efficiently and in ways that correspond to their own experience, where chatbots are one of these new solutions. One of the main problem that chatbots face today is to mimic human language, where they try to find the best answer to an input, which is not how a human conversation usually works, rather taking into account the previous messages and building onto them. Extreme programming methodology was chosen to use integrate ChatterBot, Pyside2, web scraping and Tampermonkey into Blackboard as a test case. Problems occurred with the bot and more training was needed for the bot to work perfectly, but the integration and web scraping worked, giving us a chatbot that was able to talk with. We showed the plausibility of integrating an AI bot in an educational setting. 

\end{abstract}


\section{Introduction \& Background}
\label{sec:intro}
Finding new ways to help students has always been important in the field of higher education. Innovation is a key component in trying new solutions for the students to learn efficiently and in ways that correspond to their own experience. Chatbots are one of these new solutions \cite{adamopoulou2020overview}. This work attempts to further the research in this area, by creating an educational chatbot integrated into the Blackboard platform.

Chatbots are computer programs that can hold a conversation with humans in natural language. They are used in a great variety of different fields, including for informational and educational purposes. Pedagogical chatbots have been found to have a great positive influence on students’ learning. Research shows that the interactivity of chatbots makes them a great help in creating an interactive and efficient learning environment \cite{imran2009interactive}. Our aim in this research work is to create a chatbot that could be integrated into Blackboard and could serve as a virtual teaching assistant, by being able to answer questions from students about the course to facilitate and support multimedia learning objects \cite{imran2012multimedia}.

%The paper is structured as follows. The Related Works section \ref{sec:relatedwork} details the state-of-the-art in the field of educational chatbots. Section 3 contains the methodology, while Section 4 presents the implementation details of our chatbot. Section 5 contains the results and discussion on the challenges and current limitations. Lastly, section 6 concludes the article.

%\section{Related Work}
%\label{sec:relatedwork}

In order to develop a state-of-the-art chatbot, many technologies are out there as discussed in \cite{adamopoulou2020overview}. Research work in \cite{roller2020recipes} briefly discusses the history and type of chatbots. To create a chatbot, it is also extremely helpful to have some examples of how other chatbots’ discussions are \cite{adiwardana2020towards, roller2020recipes, clarizia2018chatbot}, and how to make chatbots seem more human \cite{adiwardana2020towards} and trustable \cite{folstad2018makes}.
Even though this implementation focuses on educational bots, it is worth noting the wide spectrum of chatbot usage such as in education, health, productivity, to name a few. Chatbots can be used in almost all circumstances and fields of work or study. Starting from the health department and the victim aid bot named ‘SPeCECA’, whose job is to facilitate victims and witnesses of accidents and sudden illness. The bot aims to prevent them from inflicting avoidable damage to the patient until the trained response unit arrives at the incident \cite{ouerhani2020spececa}. Moving on to chatbots like ‘ActionBot (ABT)’ which help mitigate differences between team members and can be used in a learning or working environment. It can be used to improve team cohesion and overall performance while avoiding lack of communication and potential hindrances to productivity \cite{benke2020chatbot}. To finally chatbots like SLOWBot, one of the thousand variations of chatbots implemented with aim to aid the user in their pursuit of a better and healthier lifestyle \cite{gabrielli2018slowbot}.

This adaptability and potential use of chatbots is one of the reasons to explore existing off-the-shelf chatbots, and try to see more potential possibilities it has to offer. For instance, such a tool could be used as mode of social interaction \cite{imran2016analysis}, or to teach information security concepts \cite{nordhaug2014forensic, alawawdeh2014norwegian}. The first problem with this variety of course is that there exists and equal variety of ways to implement them, each with their own strengths and weaknesses. A quick example of the complexity and structure such a chatbot can easily reach is SPeCECA the victim support bot, given in Fig \ref{fig1}:

\begin{figure}[ht]
    \centering
    \includegraphics[width=15cm]{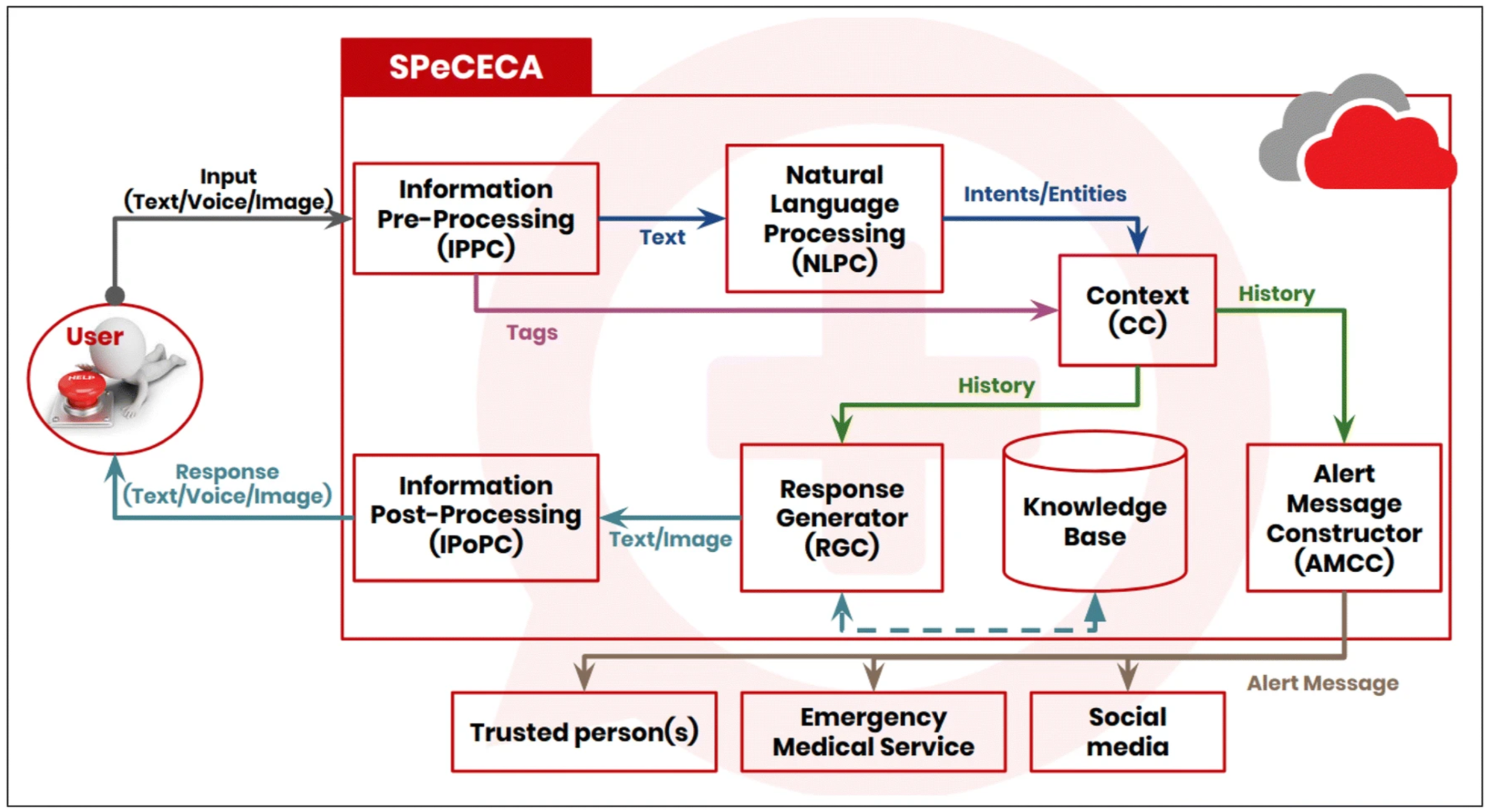}
    \caption{SPeCECA’s components and actors \cite{ouerhani2020spececa}}
    \label{fig1}
\end{figure}

Although, usually simpler structures are sufficient for the chatbot to be effective in its task.

One of the main problem that chatbots face is to mimic human language \cite{caldarini2022literature}. They try to find the best answer to an input, which is not how a human conversation usually works, rather taking into account the previous messages and building onto them.

There are many chatbots developed for various purposes. For example, chatbots used for educational purposes have been found to have a number of benefits \cite{fryer2019chatbot, hiremath2018chatbot, winkler2018unleashing}. A WhatsApp chatbot for students \cite{balatamoghna2022enhancement} helped a group chat made for students become more active during the pandemic. It had both entertainment purpose, keeping the group active with a game, and educational purpose, with an API attachement where students could ask questions about a specific topic.

Another study Gupta and Chen \cite{gupta2022supporting} tried to find how a chatbot could help create a more inclusive learning environment. It would make it easier for all students to have access to the proper resources, especially for students with learning disabilities or disorders (the possibility to hold conversations through chat or through audio for hearing or visually impaired students is a strong point for ex- ample). It would also be able to adapt to students with very different environments. And there are different types of learning that a chatbot can use for students depending on their preferences: some learn better with visual learning (videos, images, color codes), kinesthetic learning (with activities like games and quizzes for example) and auditory learning (by providing podcast resources or audio flashcard games) \cite{dalipi2017analysis, pireva2015user}.

Notable examples of those kind of chatbots are many. For instance, @dawebot is a chatbot that tutors students in a wide range of topics and courses through an assortment of well-aimed quiz questions, which helps their focus and concentration \cite{pereira2016leveraging}. CidoPoliBot is another state-of-the-art bot in development which gamifies education and takes into account a specific target audience. It is developed by Fadhil and Villafiorita and it educates children about proper lifestyle choices. The initial testing yielded very promising results showing a clear advantage of the bot over the previous cut board tutoring methods on which it is based \cite{fadhil2017adaptive}. Another significant step in educational innovation is a chatbot named ‘SingGuru’ which aids in the learning of American Sign Language. It provides direct feedback translating the teacher’s movement from sing language into detailed text. It is currently used as a tutoring aid to a course for said language and it can currently help the user orient themselves in the course but also takes an active part in the learning process and tests. Questionnaires have shown an increase in student performance of over 20 percent \cite{paudyal2020evaluating}. In addition, we must mention CapacitaBOT a fascinating bot which aids persons with intellectual disabilities. Due to the recent lock downs and the modern way of life in general it’s hard for some people that struggle with daily social interactions to practise and help them better fit into the society. This bot aids and educates them by honing their social abilities and training them for real life situations \cite{mateos2022chatbot}. Lastly, there is a chatbot called Jill Watson which is closer to what we are trying to accomplish here. This bot is a virtual teacher assistant which has successfully helped and interacted with over 700 students. JW is able to answer frequent asked questions and uploaded announcements without any problem \cite{goel2018jill}.

As we can see the potential for interactive learning with chatbots can be a very interesting and potentially useful field of study, which can hold vast potentials for improvement with new ground-breaking innovations that have started to appear in the resent years \cite{torous2021growing}. It can greatly facilitate class tutoring \cite{lee2020using}, and improve experience for both students and teachers, since they cannot always be available to answer every student questions \cite{siblini2021towards}, as our current study ties to prove. But it has the future potential of being an autonomous entity, helping the members of society that need it the most.

\section{Methodology}
\label{sec:methodology}

Our research work was guided by the double diamond design process \cite{liu2009international}:

\begin{itemize}
    \item The first phase consisted of researching chatbots and the state-of-the-art on how they work and can be employed for educational purposes. 
    \item In the second phase, we used our research to formulate a problem statement, and define the objective of this study. We decided to develop an AI chatbot to integrate into Blackboard for educational purposes. Our bot could be added to the webpage, and converse with students about the course topics. We wanted to use webscraping, so that the bot could find by itself documents about the topics on the Blackboard page, and train itself with them.
    \item Then we explored possible ways of implementing the chatbot, trying different solutions for each goal in the study and each problem that we encountered.
    \item Finally, in the last phase, the solutions we implemented started to converge towards a working prototype, as we found what worked for the chatbot and what fit the best for the objective that we settled on for this research work.
\end{itemize} 
 
This design process has allowed us to experiment with the study. In the first part of researching the topic of chatbots, we learned a lot about the current state of chatbot technology and how it is used for educational purposes. The second part let us bring new elements to the chatbot, adding new features that were not previously integrated.

\begin{figure}[ht]
    \centering
    \includegraphics[width=15cm]{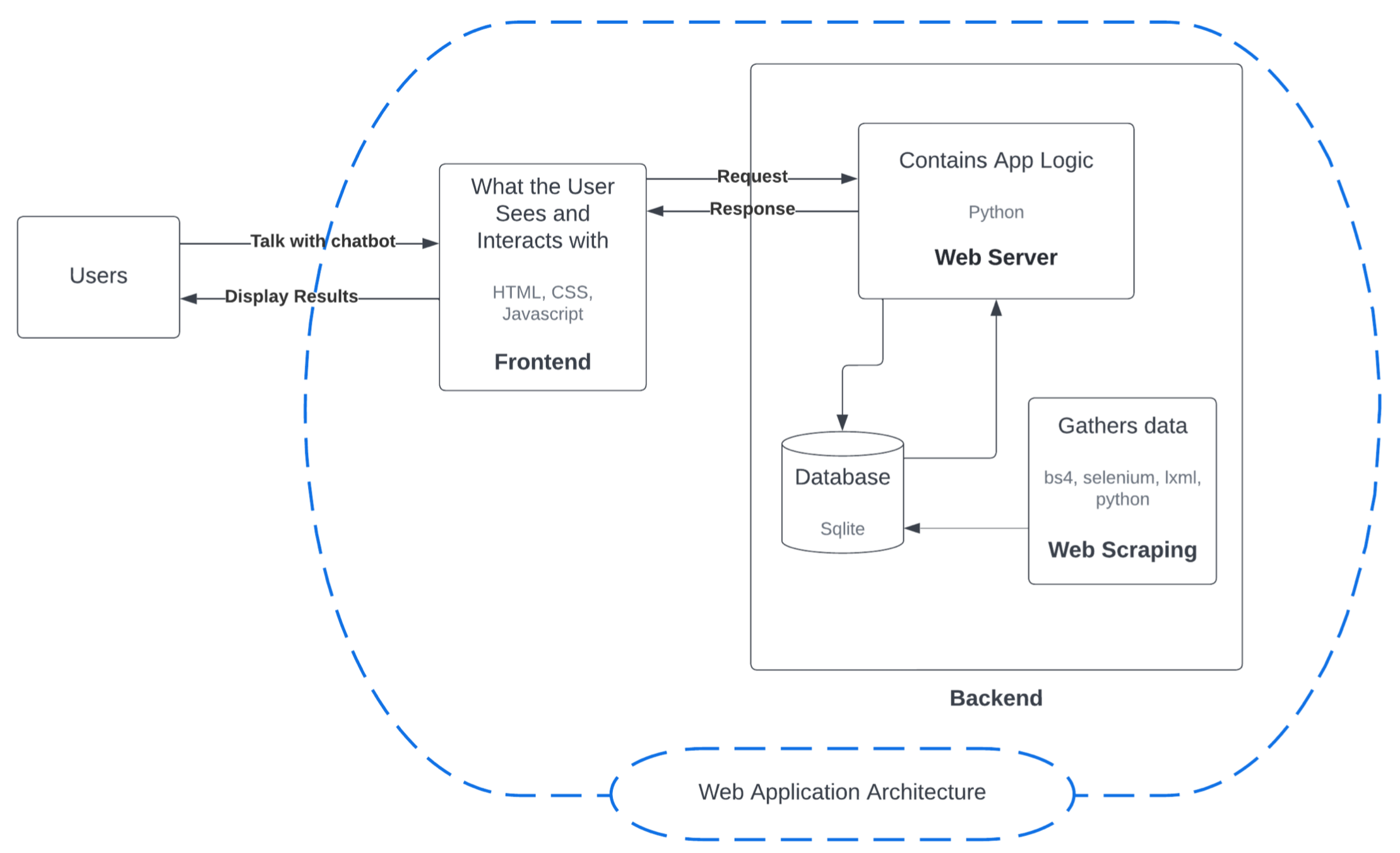}
    \caption{Architecture design of the proposed solution}
    \label{fig2}
\end{figure}

Figure \ref{fig2} shows the architecture of our proposed solution.

Python was used for coding the application. The main packages for Python in this study was ChatterBot and PySide2, the framework for the bot and the library for the graphical user interface. 

Tampermonkey and Javascript were used for the integration of the chatbot into Blackboard.
Blackboard was used to test out the bot and its web scraping. This version of Blackboard was a local instance given to us by the IT-department to test out different aspects of the project on.

\begin{figure}[ht]
    \centering
    \includegraphics[width=15cm]{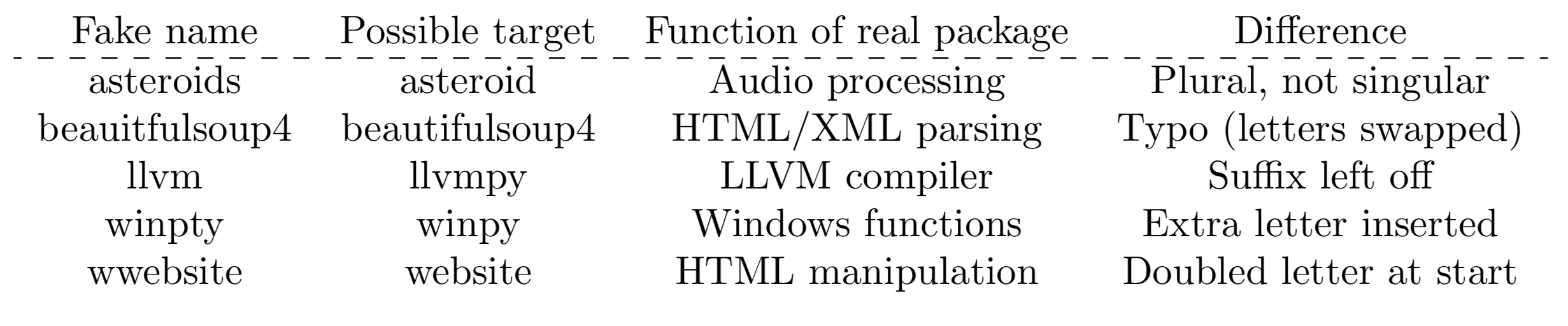}
    \caption{Package squatting \cite{ref15}}
    \label{fig3}
\end{figure}

Since the bot will be integrated into Blackboard, this means that Feide will be the main security for the chatbot. The main security concern for this research work is ”Poison packages”. By using Python, a lot of the packages used in this work can be open source. The simplicity of downloading and installing packages in Python, which has led to a range of cyber-criminal attacks against package managers. The criminal can either Trojanise a repository or make fake packages of a private project that is in use. By doing this the attacker may be able to deceive a company’s whole development team, or even the organization’s official software build system, into updating private code from an untrusted (and dangerous) external source if the company’s auto-updating processes aren’t properly safeguarded\footnote{ https://nakedsecurity.sophos.com/2021/03/07/poison-packages-supply-chain-risks -user-hits-python- community-with-4000-fake-modules/}. This sort of trick is known as a supply chain attack. Figure \ref{fig3} shows package squatting where there are packages with typos that the criminals hope people will use.

\section{Chatbot Implementation}
\label{sec:implementation}

Since the task involved developing a domain specific conversational AI bot for a Hyper Interactive Intelligent Presenter \cite{imran2014hip}, we needed to find a framework that could be used to solve the task and also a graphical user interface framework that could easily be used. Since the bot needed to be trained to converse on a domain-specific topic for a course extracted later on from the video classification framework \cite{kastrati2019integrating}, the chosen framework would need to support this. The main task for this conversational AI bot was to assist users in acquiring basic knowledge about a topic and/or for querying relevant material available on the course page. Such a chatbot can be a great way to acquire students opinions and sentiments towards certain aspects related to the teaching activities as well \cite{edalati2021potential,10.1145}. It can also be a great way to observe students' learning experience \cite{9759714} and behaviour \cite{patel2019combating, pireva2019evaluating, tran2020understanding} when the interact with the course through this bot and to measure their engagement in a game-based learning approach \cite{nuci2021game} or to learn a dropout pattern \cite{lim2021design,imran2019predicting}.

\begin{figure}[ht]
    \centering
    \includegraphics[width=10cm]{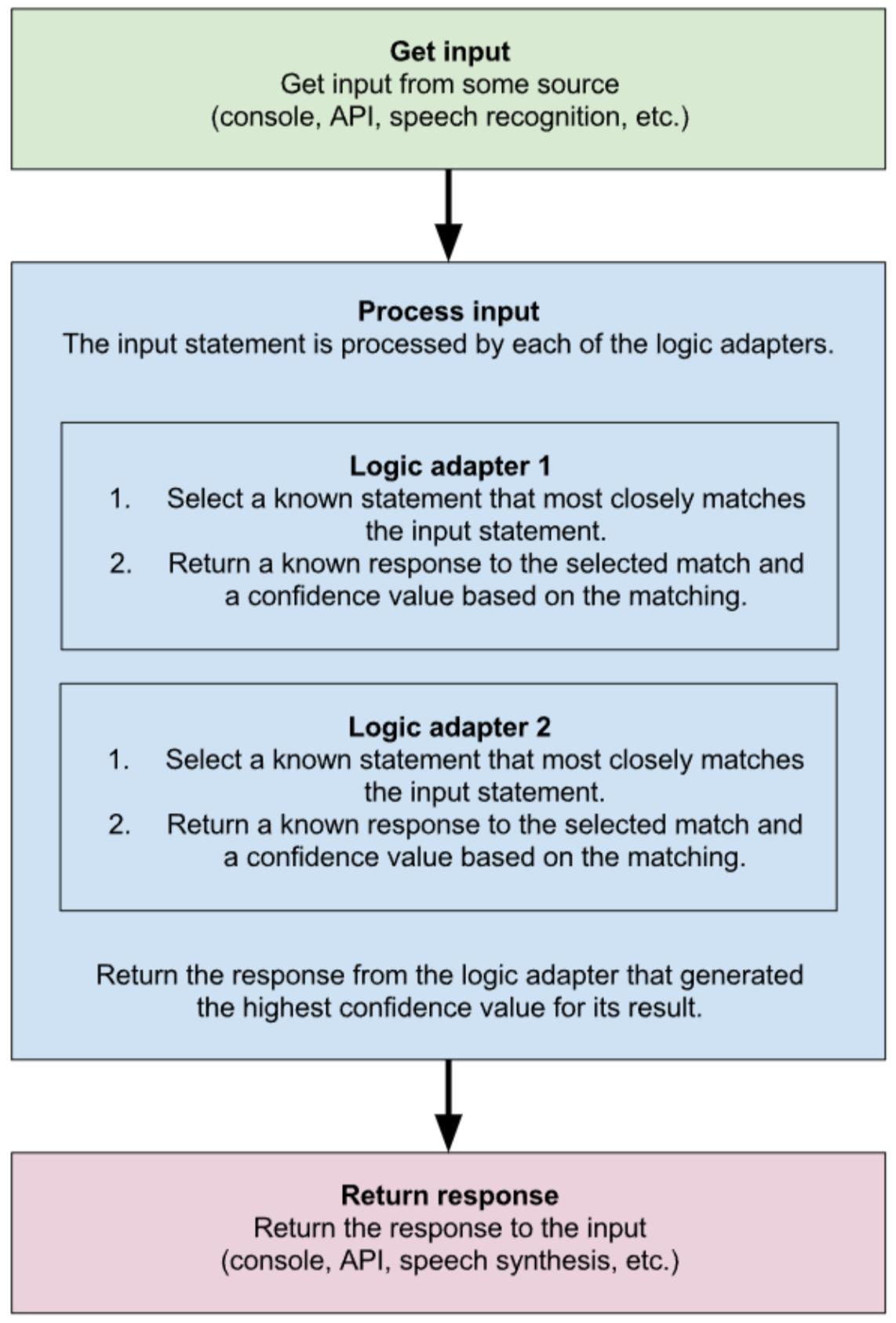}
    \caption{Process flow diagram}
    \label{fig4}
\end{figure}

ChatterBot was the framework that was chosen to be used for this article. ChatterBot is a Python library that makes it easy to generate automated responses to a user’s input, and uses a selection of machine learning algorithms to produce different types of responses \cite{ref7}. In figure \ref{fig4} you can see how the ChatterBot framework works, from getting the input and choosing what to output\footnote{https://chatterbot.readthedocs.io/en/stable/}. 

The ChatterBot framework allows you to change a lot for how the bot will process the input statement or how the bot will logically choose an appropriate output statement. There is also a way to have a default response, but this is not something we have implemented. Under is a small list of the main things you can change with the framework. If you do not specify these parameters when you set up the chatbot, the framework will choose the parameters for you. The list also gives a short explanation on what each item is.

\begin{enumerate}
    \item Preprocessors: simple functions that modify the given input that the conversational bot receives before it gets processed by the logic adapter
    \item Logic Adapters: determines the logic for how the bot selects a response to a given input
    \item Storage Adapters: provides an interface that allows the bot to connect to different storage technologies
    \item Filters: efficient way to create queries that can be passed to the storage adapters used by the bot. They reduce the number of statements the bot has to process before giving a response

\end{enumerate}

\begin{figure}[ht]
    \centering
    \includegraphics[width=16.5cm]{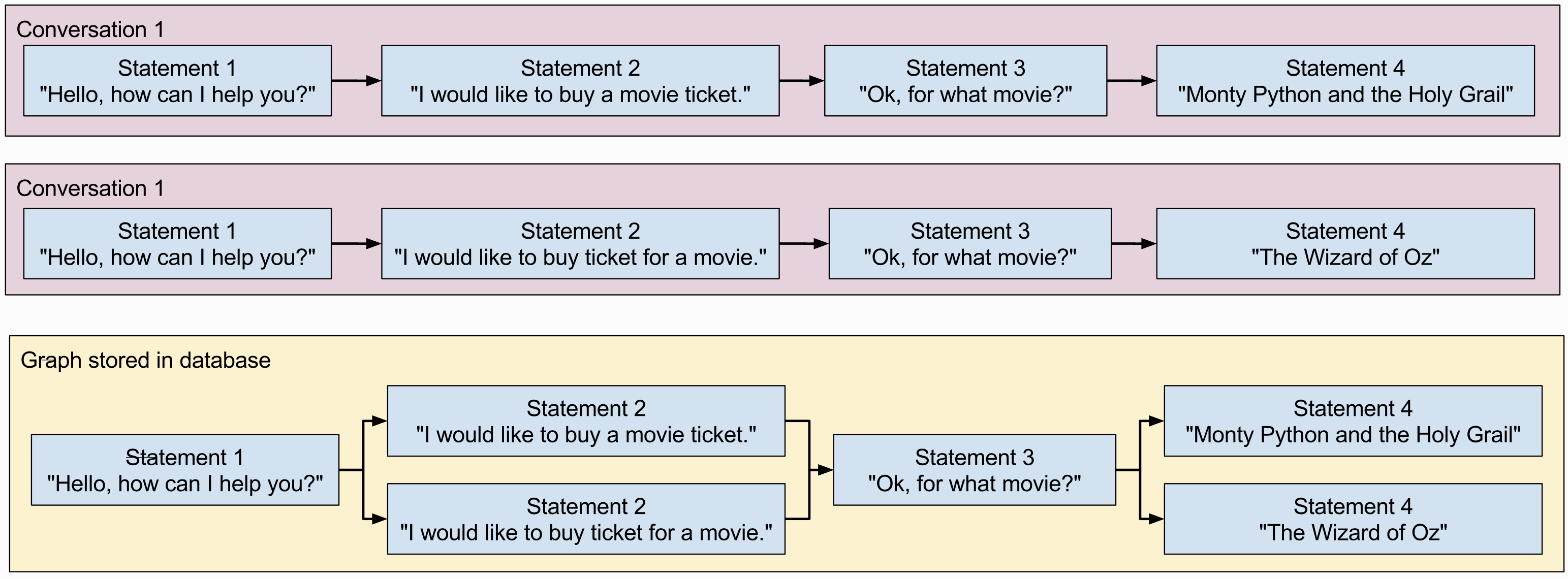}
    \caption{Conversations in database}
    \label{fig5}
\end{figure}

The most important parameter for this study was the training and what type of training to be used for the bot. The ChatterBot framework includes tools that help simplify the process of training a chatbot instance. The training process involves loading example dialog into the database used by the bot. The graph data structure that reflects the sets of known statements and answers is either created or built upon. When a chatbot trainer receives a data set, it constructs the necessary entries in the conversational bot’s knowledge graph to accurately reflect the statement inputs and responses. Figure \ref{fig5} shows how conversations with the bot are stored as graphs in the database\footnote{https://chatterbot.readthedocs.io/en/stable/training.html}.

The main ways to train the bot is using a list of statements where the order of each statement is based on its placement in a given conversation, training it with corpus data or training it with the Ubuntu dialog corpus \cite{lowe2015ubuntu}. For this research work the English corpus dialog\footnote{ https://github.com/gunthercox/chatterbot-corpus} set was chosen. This trained the bot so that we could easily communicate with it and the bot could give appropriate outputs to out inputs. Each time the bot have a conversation or "learn" something new, it gets stored in a database. In listing 1 you can see a simple conversational bot that uses the terminal to talk with the user and is trained on the English corpus data. %After a bunch of inputs and training data, the learning feature can be disabled by changing what is shown i listing 2

\begin{figure}[ht]
    \centering
    \includegraphics[width=16.5cm]{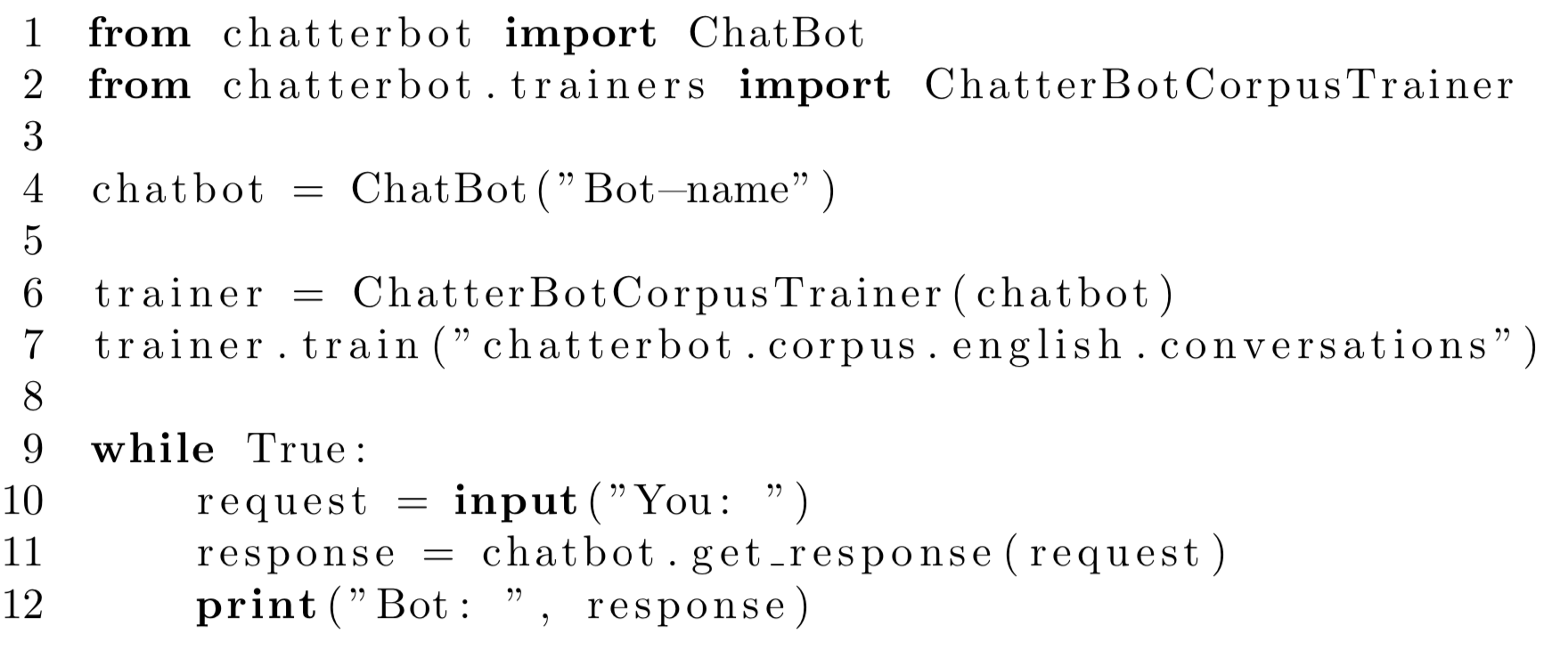}
    \caption{Listing 1: Terminal example for the bot}
    \label{fig6}
\end{figure}

The framework that was chosen was PySide2 for GUI. PySide2 is a Python binding of the cross-platform GUI toolkit Qt and is one of the alternatives to the standard library package Tkinter for Python. There is two different library package for the GUI toolkit Qt. PySide2 and PyQt5. Even though there are two packages, they are basically the same. They are 99\% the same\footnote{https://www.pythonguis.com/faq/pyqt5-vs-pyside2/}. The reason why there is two packages wrapping the same library, is because of licensing. PySide2 uses a LGPL license while PyQt5 uses GPL or commercial license\footnote{https://www.pythonguis.com/faq/pyqt-vs-pyside/}. Since the libraries are the same, it is easy to take a PyQt5 project and use it with Pyside2.

\begin{figure}[ht]
    \centering
    \includegraphics[width=16.5cm]{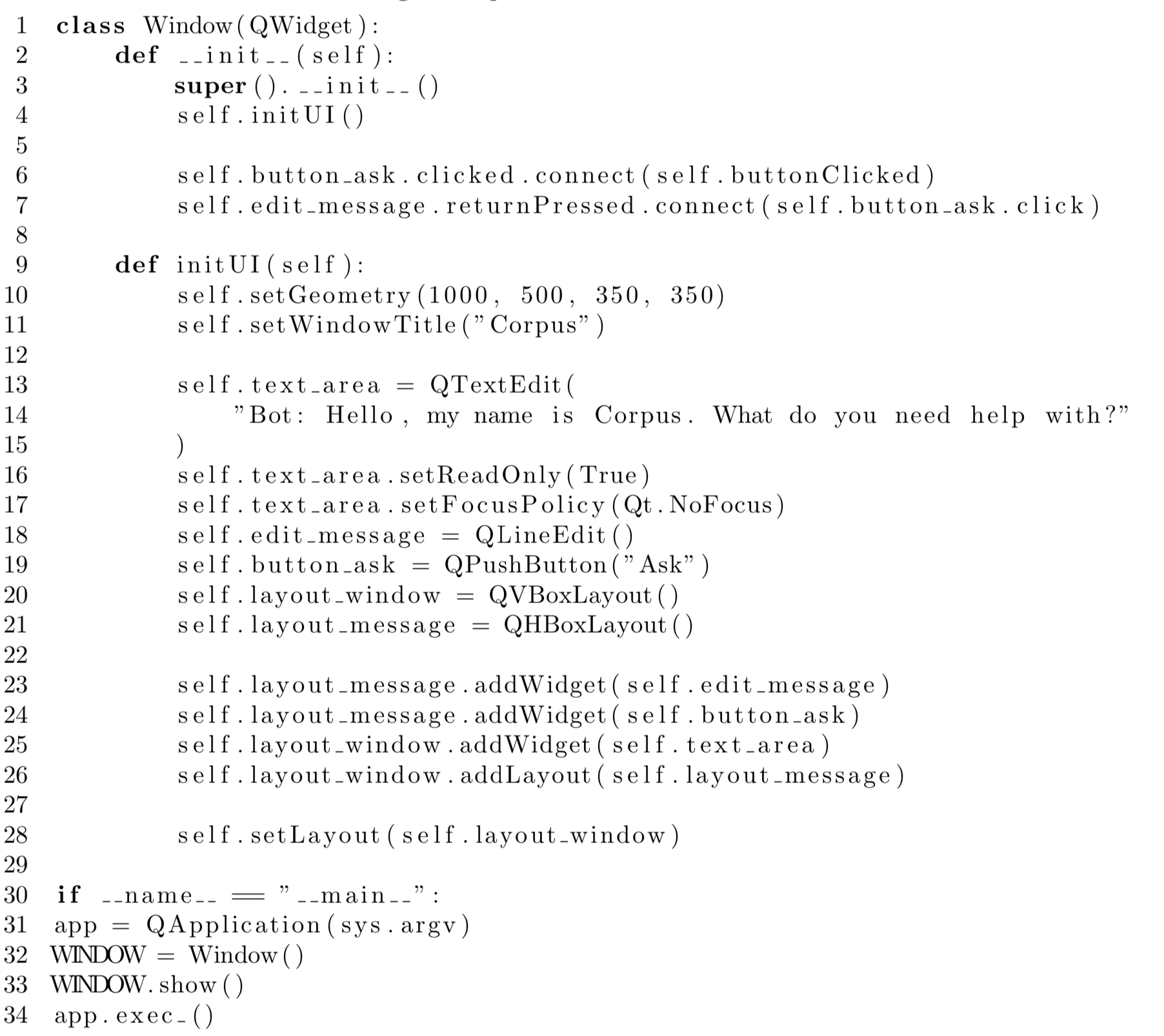}
    \caption{Listing 2: Graphical user interface for chatbot}
    \label{fig7}
\end{figure}

In listing 2 you can see the main code for setting up and running the graphical user interface. The interface has a text area for displaying the conversation, a input field to ask the question and a button to press to ask the bot. The reason why the button is there and not just using "enter", is because some people may not know you can use "enter" to ask questions. Layouts are made to keep the different widgets and to make it a bit prettier for the eyes to look at. The text area is made to be read only. This is so that the user cannot write in it and the area is only used to display the conversation between the bot and the user.

\begin{figure}[ht]
    \centering
    \includegraphics[width=12cm]{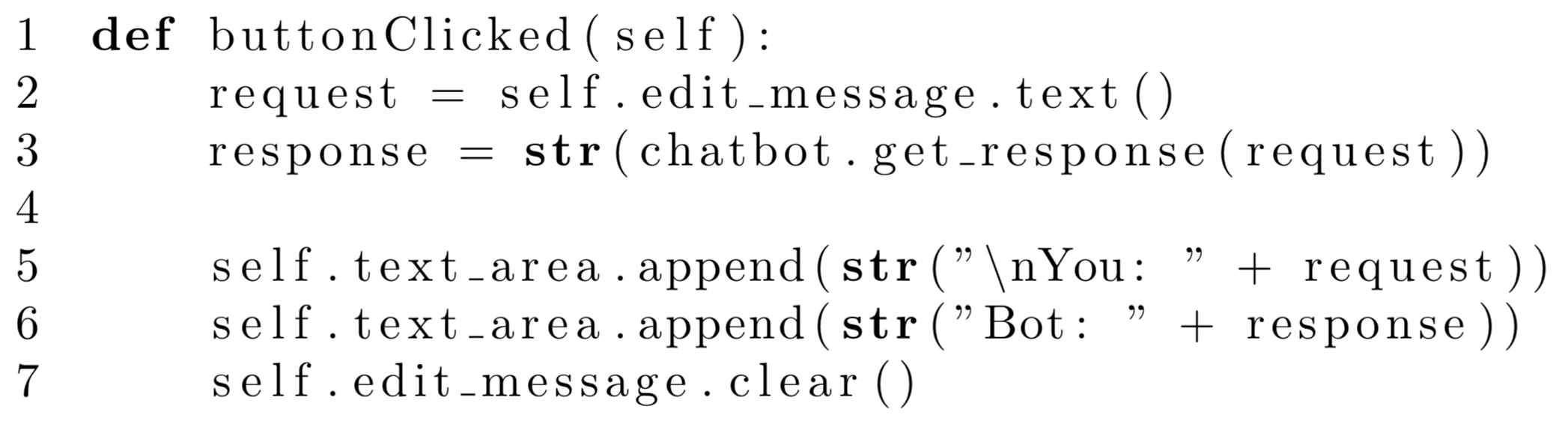}
    \caption{Listing 3: Request/response for the bot}
    \label{fig8}
\end{figure}

Listing 3 is the main connection between the ChatterBot framework, mentioned in the beginning of this article, and the graphical user interface. The request from the user is "taken" from the input field and is then sent to the bot which then gives back a response. This happens when either the user presses "enter" after a question or presses on the Ask-button. Line number 6 and 7 in listing 3 shows how the button-function is connected to run when the button is pressed or ”enter” is pressed. After the bot has given back a response, the request and the response is then displayed on the text area. This displaying is done with each new conversation having a newline before itself, this to make it easier to read. There is also added to strings in front of the response and request. You in front of the request and Bot in front of the response. After they are displayed on the text area, the input field is cleared.

\subsection{Database Management}
\label{sec:db}

The database that Chatterbot creates consists of three tables. One table contains tags that rep- resent different topics that the bot has been trained on. The main table contains all the training data with timestamps and what the statements are a response to. If a statement is not part of the training the bot has done, it will have a field called "conversation" be blank, with it as training if the statement is part of the training. The final table is connection table between the training data and the topic tags.

The program for merging the databases will start by finding all the databases of the type .sqlite3, which is the format Chatterbot saves the database file in. If there is one or less databases in the folder the program is in, it will tell the user that there aren’t enough databases to merge, as well as which database type the program supports. It will then ask the user which database they want to merge into. Once a database is selected, the program will start the merging process by first trimming down all the other databases to the statements that has a blank conversation field. It will then loop through the databases and add any statement that does not exist in the selected database to add the statement to the database as shown in listing 4. A statement is defined as not existing in the selected database by comparing all the fields except for the id field. Once all the databases has been merged into the selected database, the program will tell the user which database they had chosen to merge into, in case the user forgot their selection. The program is ignoring the training fields as they should stay the same through all the different databases, and if there is one person that decided to add some training to their database, the user can just select the database with the updated training set to merge into.

\begin{figure}[ht]
    \centering
    \includegraphics[width=16.5cm]{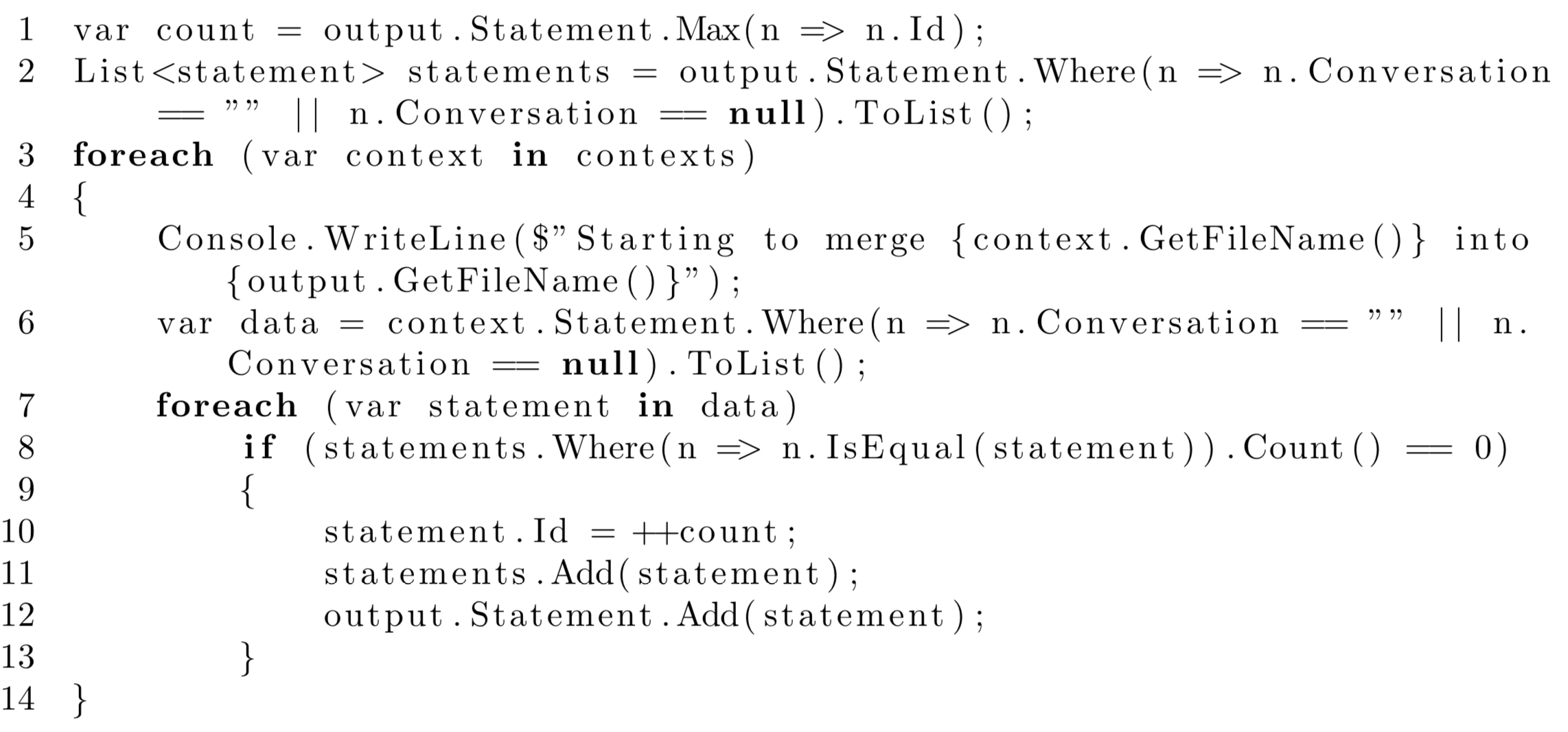}
    \caption{Listing 4: Database merging}
    \label{fig9}
\end{figure}

\subsection{Web scraping}
\label{sec:ws}

Since our state-of-the-art chatbot was meant for aiding students and instructors in their respective courses, the blackboard web page was an obvious way to try and obtain data to train it. And we can scrap the information directly as well from course contents available on blackboard \cite{imran2011blackboard}. To do this we attempted to get all the necessary information with a known technique called web scraping. ‘Web scraping is used to transform unstructured data on the web into structured data that can be stored and analysed in a central local database or spreadsheet’ \cite{sirisuriya2015comparative}. That way we aimed to get data from the information pages of each course and any wiki or site they might site and train the chat-bot to answer questions based on them. That would ease the students and make the data easier accessible, since they wouldn’t have to spend time going through everything or even try to contact their supervisor in case, they had trouble finding uploaded data.

There were many good choices of tools like HTQL, iMacros and IRobotSoft each one of them a top choice with unique features \cite{giannini2014web}. Nonetheless, we thought the best course of action was to try and implement our own to get better acquainted with the methodology. To do this, we looked at a wide variety of web scraping techniques and programs and after discussion we decided to go for selenium and beautiful soup python libraries. The main reason we choose them was their popularity and ease of use, so that we might obtain fast and easily the information we needed. After some initial trial and error, we got a fully working web scraper up and running in relatively short time. This first attempt was able to collect a wide range of data, which we managed to categorize, from some test websites that didn’t require log in. The next step was password protected websites, this part was a bit harder to fully implement but again we managed to set up a program that could log in and then scrape some sites provided by the tutorials.

\subsection{Integration}
\label{sec:int}

Since we were unable to make any direct changes to blackboard, we had to find another way to integrate our chatbot into Blackboard. A solution we found for this which we decided to make use of is Tampermonkey\footnote{https://www.tampermonkey.net/} which is an add-on for browsers that allows for the use of userscripts. These userscripts can both add and remove functionality from websites, and we made use of both parts in our work.
The script is made up of a small amount of functions. We started by creating a button that will be placed in the lower right of the screen. Clicking the button will open a chatwindow where the button was placed and remove the button. The chatwindow consists of four parts, a gray input box for the user to talk with the chatbot with a button to the right of it. Clicking on the button or pressing enter will then send a request to the chatbot with the question the user wants to ask the chatbot.

The question will then be added to the white box above the input field and once the response from the bot comes back, it too will be added to the white box. Above the white box, there is another button that will close the chatwindow and add back the first button. The script will remember the discussion as long as the page isn’t reloaded or changed and will thus still be there if the chatwindow is closed and reopened again. Such a tool could easily be integrated into other pedagogical platforms \cite{imran2016automatic}.

\section{Results \& Discussion}
\label{sec:results}

\begin{figure}[ht]
    \centering
    \includegraphics[width=16.5cm]{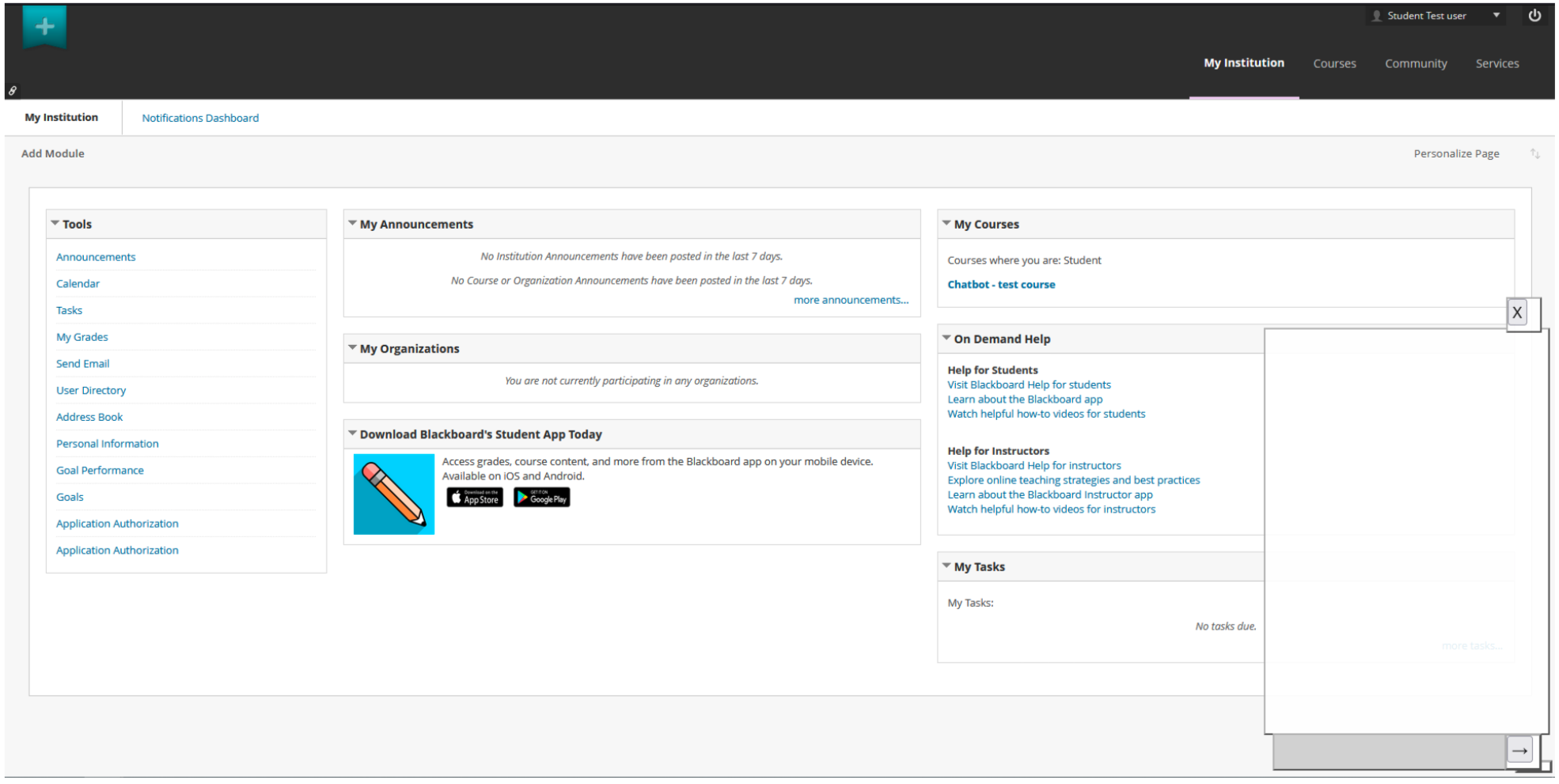}
    \caption{Full Blackboard page with bot in the corner}
    \label{fig10}
\end{figure}

With all the implementation mentioned above, the Blackboard page looks like figure \ref{fig10}. The figure shows how the chatbot is integrated into Blackboard. The grey part under the white box is the input area, while the white box is the conversation area. If the user asks question, the bot will answer it.

The next figure, figure \ref{fig11}, is showing how a conversation with the chatbot looks like. As previously mentioned, the questions/answer has a help-string in front of each. This is to help know who asked and answered what. As shown in the figure, the bot does not always know what to answer. Sometimes the bot also answers something that does not have any correlation with the question. The easiest fix for this s simply more training for the bot. With more training, the bot will get better and have more confidence giving the right answers. More training equals better answers.

\begin{figure}[ht]
    \centering
    \includegraphics[width=4cm]{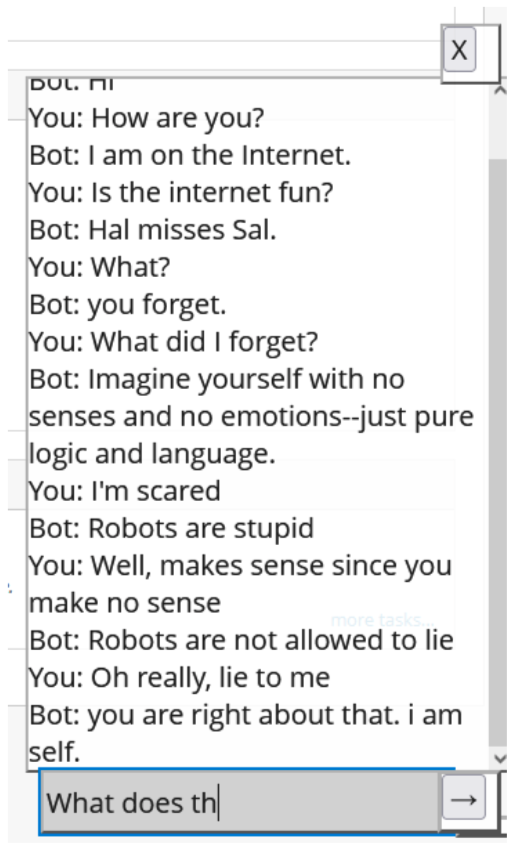}
    \caption{Discussion between user and bot}
    \label{fig11}
\end{figure}

The framework stores each conversation as a graph in the database. Each time the user asks, the bot stores it into the database. If the bot ask "How are you doing?", and the user answers "Doing fine", "fine" or "good", all these will be stored as answers to the question in the database. Then the bot know what to answer if the user asks the same question. "Hi there" and "How are you doing?" is stored a lot because that was always the first questions the authors asked during development of the bot. That means the bot knows a lot on how to respond to those questions and inputs.

\subsection{Challenges}

\textbf{installation}
We had quite a lot of problems when it came to installing Chatterbot and making it work. We were all able to install version 1.0.4 of Chatterbot, but not the newest version of Chatterbot, version 1.1.0. We were looking for what the cause of this could be for a long time. What we did notice was that package called spacy did not want to install newer versions, and that it tried to install older and older versions of it until it was able to install a version compatible with Chatterbot version 1.0.4. With no idea on how to fix this, we decided to use version 1.0.4 of Chatterbot as it was the latest available version that we were able to install, but it wasn’t without its own problems. Version 1.0.4 of Chatterbot was made for a python version of 3.7 or lower, but we were using newer python versions which had removed or changed two functions that Chatterbot was dependent on. Due to this, we had to go into the packages we had installed and find the function that was removed or changed and then replace it with the newer method. The functions we had to change was "time.clock()" to "time.perf counter()" in sqlacademys util/compat.py and "collections.Hashable" to "collections.abc.Hashable" in yamls constructor.py. With this done, we were able to get Chatterbot up and running.

\textbf{design aspects}
Now the bot does not have any design behind it, it is just a simple chatbot that can be interacted with. As you can see in figure \ref{fig10} it does not look like it belongs there, and it also does not look nice. One of the things that could have been made, is a small bot-button as you can see on other pages that have chatbots. Pushing this button "summons" the bot conversation on the page, almost like opening a message. Also not having such hard corners for the bot. Right now it has very sharp corners, but would be nicer to have rounder corners. The size is also something that would need to be changed design wise. All these different design changes should be talked about with someone having some design background, since most of the programmers do not have the same knowledge as they do.

\textbf{training}
The bot is trained on two different methods: on a list of responses and questions, and on a corpus dialog set. This makes it so that the bot can talk, but still needs a lot of training to be more human like. We initially tried to train the bot using the Ubuntu Dialog Corpus but after a few hours we turned off the training because it was said training with this data set could take as much as a week to finish. Training the bot on large data sets will help it a lot with the learning part, which in part will help the users when talking with them. This large and time consuming training can be done by using/renting a virtual machine and letting the chatbot train there. After the training is done, the database file will then be used with the local chatbot. This can also be made into two different instances of the chatbot. One that only trains and the other one for the talking. Then you can just continuously feed the learning bot with material. Using the database merger mentioned in 4.2 can be used for combining the two databases after each training period.

\textbf{Web scraping}
The main way from which we planned to add data is the web scraping of the blackboard website, the course information page specifically seemed to be a good start. Although the web scrapper seemed efficient when collecting data from other websites we weren’t that successful with blackboard. Unfortunately, we didn’t take into account Feide’s security system which prevented us from taking the necessary data to train the bot efficiently. Nevertheless, that doesn’t mean that it cannot be of use as it can always collect data from provided supportive material, like Wikipedia or other websites, each professor might provide to his students. Another important topic is that we didn’t implement our own training class as the chatbot framework we used allows. The reason for this is that the web scraper wasn’t set up the right way for training, as the data format wasn’t .yml or a list.

\textbf{integration}
Due to only getting access to a test-environment of Blackboard that we couldn’t directly change the source code of, we were unable to directly integrate the chatbot into Blackboard. It would have been a lot better to implement the chatbot into Blackboard, but due to this problem, we had to rely on making an external modification of the website. While Tampermonkey works for modifying blackboard, we can’t look away from the fact that we are not only relying on a third party tool for this but also in a sense making a third party method for adding the chatbot to Blackboard.

\subsection{Limitations \& Improvements}
\label{sec:lim}
One of the limitations of this chatbot is its learning ability to understand context and use semantics \cite{qaffas2019improvement, kastrati2019impact, virkar2019humanizing}. The bot can be made to comprehend the semantics either using ontology, \cite{kastrati2016semcon, hallili2014toward, kastrati2015analysis, spiliotopoulos2020semantics, kastrati2015semcon, al2011ontbot}, or by incorporating concept vector space models \cite{kastrati2019performance, kastrati2015improved}, or by utilizing embeddings \cite{bensalah2020combining, kastrati2020wet, gamage2020impact, kastrati2019integrating}, to enhance the personalized learning trajectory thus making the chatbot works like a personal learning companion. In addition to teaching and learning activities, the chatbots can also contribute to the administrative part of the education sector by providing students with guidance and assistance related to admission process, tuition fee, etc, \cite{BHUTORIA2022100068}. Apart from this, the bot can further be trained to understand domain-specific jargons or be able to converse in multi-languages.  

\section{Conclusion}
\label{sec:conc}
The main goal of this research work was to create an educational chatbot. We made a chatbot that is integrated into Blackboard and can be trained on a specific topic so that it can answer student’s questions about it. We also researched webscraping and tried to use it on Blackboard so that the bot could learn from the Learning Materials directly on the course page, but we encountered some problems specific to Blackboard that complicated this task. The webscraping part is still working on lower-security websites and can be exploited for the chatbot. For the future works related to this study, the first thing to do would be to connect the webscraping part to the chatbot, as they are current in completely separate files and not connected in any way. The data gathered by the webscaper is not directly usable by the bot, so implementing that is necessary for the interaction. Then the chatbot should be trained, which can be done through webscaping on webpages other than blackboard, with trusted webpages given by the professors for example, or simply with \textit{.yml} files made specifically for the chatbot.

\bibliography{access.bib}{}
\bibliographystyle{IEEEtran}

\end{document}